# Using k-nearest neighbors to construct cancelable minutiae templates

**Qinghai Gao,** School of Engineering Technology**,** SUNY at Farmingdale, NY 11735, USA
Email: Qinghai.Gao@farmingdale.edu

**ABSTRACT**

Fingerprint is widely used in a variety of applications. Security measures have to be taken to protect the privacy of fingerprint data. Cancelable biometrics is proposed as an effective mechanism of using and protecting biometrics. In this paper we propose a new method of constructing cancelable fingerprint template by combining real template with synthetic template. Specifically, each user is given one synthetic minutia template generated with random number generator. Every minutia point from the real template is individually thrown into the synthetic template, from which its k-nearest neighbors are found. The verification template is constructed by combining an arbitrary set of the k-nearest neighbors. To prove the validity of the scheme, testing is carried out on three databases. The results show that the constructed templates satisfy the requirements of cancelable biometrics.

## 1. Introduction

As one of the most popular biometrics, fingerprint has a long history of usage in crime investigation. In modern days many fingerprint databases have been built and are utilized in a variety of commercial applications. These databases become the targets of hackers. Compared to non-biometric information, biometric information cannot be replaced easily. If a fingerprint database is stolen, the fingerprint templates could be used to conduct identity theft or impersonation. Therefore, it is extremely important to protect the database. However, it is a challenging problem to protect biometric data with existing security techniques. Unlike password, biometric data cannot be secured with cryptographic one-way hash function due to the non-exact reproducibility problem. Biometric data can be encrypted. However, decryption is required upon recognition, during which it can be stolen.

Ideally, it is desirable to design a security mechanism that can easily revoke a compromised template and reissue a new template based on the same biometrics – cancelable biometrics. According to [1], cancelable biometrics must meet the following four criteria:
- Diversity: distinctive templates can be generated from the same biometric data.

- Revocability: if a template is compromised, it can be replaced with a new one based on the same biometric data.
- Non-reversibility: it is infeasible to recover the original template given a transformed template.
- Accuracy: matching with transformed template does not lower recognition accuracy.

The rest of the paper is organized as the following. Section 2 briefly reviews literature; Section 3 introduces the procedures and steps of synthesizing templates, generating the verification templates, and matching templates. In Section 4, we give the results of matching. Section 5 summarizes the paper and points out possible future research direction.

## 2. Related works

In recent years, numerous researchers have approached the problem of generating cancelable biometrics. The key to cancelable biometrics lies in template transformation. Ratha et al. [2] proposed three techniques to transform biometric template, including image morphing, block scrambling, and domain mapping. Ratha et al. [3] [4] proposed generating cancelable minutiae templates with the following three different transformations. In the Cartesian transformation, the minutiae space is divided into size-fixed and sequentially numbered cells. In the polar transformation, the coordinate space is divided into sequentially numbered polar sectors. The process of transformation is done by changing the positions of the cells in Cartesian transformation and sectors in polar transformation. In the surface folding transformation, certain locally smooth but globally not smooth functions (ex., the vector function for an electric potential field) are applied to change minutiae positions.

Moon et al. [5] proposed a template protection method by using a set of secret-key defined rectangles to move the minutiae points. Wong et al. [6-8] developed a so-called multi-line code for transforming and securing minutiae template. Jin et al. [9] proposed a template protection technique by transforming minutiae template through minutiae vicinity decomposition and random projection. Lee et al. [10] proposed using its local orientation information to rotate and translate each minutia. Chikkerur et al. [11] transformed each minutia using its neighboring local texture. Zhang et al. [12] transformed minutia using its Cylinder Code - a local structure based representation.

Yang et al. [13] proposed approach of generating cancelable fingerprint templates by transforming the derivative features of pair-wised minutiae since the local features such as distance and relative angles between two minutiae are more robust against distortion. References [14-18] also utilized pair-wised minutiae to generate new features.

Tulyakov et al. [19] and Li [20] proposed a scheme for transforming biometric templates based on minutiae triplets. Sandhya [21] [22] proposed an approach of generating cancelable fingerprint templates by deriving new features from the Delaunay triangles formed by minutiae triplets. Similar schemes are proposed with Delaunay quadrangle [23] [24], pentangle [25], and hexangle [26]. Sandhya and Prasad [27] proposed a template protection method by transforming a structure formed by each minutia and its k-nearest neighbors.

Ang et al. [28] proposed a key-based cancelable template generation method, in which the key is used to reposition minutiae.

Teoh et al. [29] proposed a cancelable template generation method named biohashing, in which the template is represented with a fixed length vector. A user-specific tokenized random number is used to generate a random base containing orthonormal pseudo-random vectors, and the inner products of the template vector with every pseudo-random vector are binarized based on a threshold to generate the transformed template. Belguechi et al. [30] proposed a similar approach.

As pointed out by Nagar et al. [31], there are two drawbacks with biohashing when the key is known to the adversary: the matching performance degrades due to the quantization of features and dimensionality reduction, and the original template can be recovered.

## 3. Proposed scheme

In this paper we propose a novel approach of constructing cancelable fingerprint minutiae template by combining a synthetic template with a user's real template, as shown in Fig. 1. Specifically, each user is given a synthesized template (ST). Every minutia point from the user's real template (RT) is individually thrown into the ST, from which its nearest neighbors are found and collected. The verification template (VT) is constructed by simply combining a set of neighbors. Note that VT is a subset of ST (i.e., $VT \subset ST$), which consists of randomly generated minutiae points and there is no common minutia between RT and VT (i.e., $RT \cap VT = \varnothing$).

The verification template constructed with this approach satisfies the four requirements of cancelable biometrics:
- Diversity can be achieved by using different ST;
- Revocability can be achieved by changing ST;
- Non-reversibility: VT does not contain any real minutia. It only contains randomly generated minutiae.
- Accuracy: our testing results with three databases show that the verification templates can produce higher matching scores than the corresponding real template.

In this paper we use the k-nearest neighbors when mapping real minutiae to synthetic (random) minutiae. Since inter-class variations have been taken into account by using strong random number generator to generate the different synthetic templates for different users, we believe other minutiae mapping methods may also be utilized as long as they preserve the intra-class variations in the real template. Future research will be conducted in this direction.

Since a user's VT contains only randomly generated minutia points, we anticipate that this approach may not suffer from the record multiplicity attack, as described by Li and

Hu [32], if the following two conditions are met when constructing multiple VT for a user:
(1) Use different synthetic templates generated with strong random number generators;
(2) Use different nearest neighbors when selecting minutiae for the verification template.

For example, the 1st nearest neighbors from ST1 are chosen to construct VT1, the 6th nearest neighbors from ST2 are chosen to construct VT2, and the 11th nearest neighbors from ST3 are chosen to construct VT3. Then the random minutiae in VT1, VT2, and VT3 will not reveal much information about the real minutiae they associated with.

Specifically, three numbers ($x$, $y$, and $\theta$) associated with every minutia are independently generated with PRNG. The ranges of the two coordinates ($x$, $y$) are given based on the original dimensions of the fingerprint image. The orientation angle is set between 0 and 360. A synthesized fingerprint template consists of $N$ synthetic minutiae, as shown in Fig. 1.

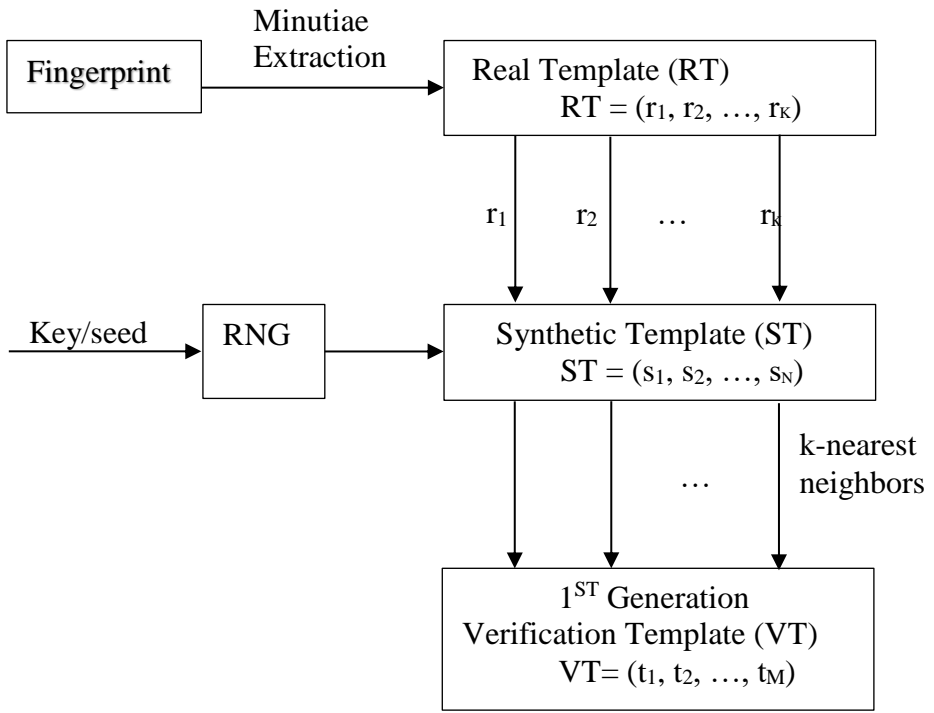

Fig. 1 Schematic diagram of transforming minutiae template ($N > M$, Random Number Generator). Note that the acronyms (RT, ST, and VT) are used as singular or plural, depending on the context.

In this paper the fingerprint templates are matched with the *Bozorth* algorithm as given by Wilson et al. [33]. It consists of three major steps:

•Transform a minutiae template into an intra-fingerprint minutia comparison table (CT)
In a template, for each pair of minutiae, a five-variable entry including the length ($d$) of the line segment between the two minutiae ($\alpha 1$, $\alpha 2$) and the angles ($\beta 1$, $\beta 2$) between each minutia and the line connecting the two minutiae, is made into a CT.
Two CTs are constructed from a pair of to-be-matched fingerprints. One is called the registered CT and the other is called the query CT.

•Construct inter-fingerprint minutia compatibility tables
Compare each entry in the registered fingerprint's CT to the entries in the query fingerprint's CT. If the differences ($\Delta d$) between the lengths of the two segments and those of the corresponding angles ($\Delta \beta 1$, $\Delta \beta 2$) are smaller than or equal to the predefined thresholds, an entry representing the two pairs of minutiae will be made into the inter-fingerprint minutia compatibility table.

•Traverse the inter-fingerprint compatibility tables to obtain a matching score between the two fingerprints.

More details about the *Bozorth* matching algorithm can be obtained from NIST Biometric Image Software [34]. All the matching results in this paper are obtained with this algorithm.

In the field of biometrics recognition, the performance of a system is often characterized with False Matching Rate (FMR) and False Nonmatch Rate (FNMR) or Equal Error Rate (EER) by selecting arbitrary thresholds. Instead of using these rates to evaluate the performance of the proposed approach, it is sufficient to use the database-based average matching scores (together with standard deviations) to compare the accuracy before and after template transformation. Higher the average score, higher the accuracy is.

## 4. Experimental results
We carried out experiments with three publicly available databases (listed in Table 1) to test the feasibility of our proposed approach. Upon designing the experiments, we attempt to answer the following questions:
(1) How does the size $N$ of the synthetic templates affect matching performance?
(2) How do different sets of nearest neighbors affect matching performance?
(3) Does the approach result in false match?
(4) Can the proposed approach survive multi-generation transformation?

The results are given below.

Table 1. Database information

| Source | FVC2004 | FVC2006 | PolyU |
|---|---|---|---|
| Database | DB1 (DB1A+DB1B) | DB2_A | DBII |
| Reference | [35] | [36] | [37] |
| Sensor type | Optical | Optical | Optical |
| Image size | 640x480 | 400x560 | 640x480 |
| Resolution | 500dpi | 569 dpi | 1200dpi |
| Dimensions of ST | 640x480x360 | 400x560x360 | 640x480x360 |
| No. Fingers | 110 | 140 | 148 |
| Images/per finger | 8 | 12 | 10 |
| Imaging sessions | 3 | 1 | 2 |
| Avg. minutiae count | 58 ± 18 | 121 ± 27 | 136 ± 47 |
| Avg, matching score* | 56.9 ± 44.6 | 45.8 ± 29.9 | 71.2 ± 54.2 |

*Matching score is obtained by matching two different templates from the same finger.

### 4.1 DB1 from FVC2004

The DB1 contains 880 real templates obtained from 110 fingers. One synthetic template is paired with each of the eight real templates pertaining to a finger, which gives eight synthetic-real pairs for each finger. For each pair, six different verification templates are constructed with the $1^{ST}$, $2^{ND}$, $3^{RD}$, $4^{TH}$, $5^{TH}$, and $6^{TH}$ nearest neighbor, respectively. Matching is carried out only between two verification templates constructed with the same nearest neighbor. Average scores and standard deviations are calculated based on the entire database. The results are plotted in Fig. 2, from which we can see that the maximal average matching score is obtained when *N* lies between 100 and 200.

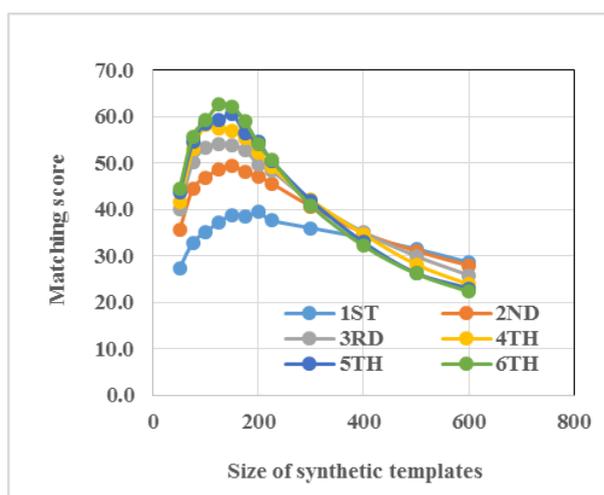

Fig. 2 Average matching score with DB1

Table 2 gives the matching results between RT and VT. The low average scores indicate the verification templates are significantly different from their real templates.

Table 2. RT vs. VT (N=200)

|  | AVG | STD |
|---|---|---|
| 1ST | 1.33 | 1.65 |
| 2ND | 1.73 | 1.69 |
| 3RD | 1.90 | 1.72 |
| 4TH | 2.04 | 1.71 |
| 5TH | 2.16 | 1.73 |
| 6TH | 2.26 | 1.68 |

4.2 DB2_A from FVC2006

The DB2_A contains 1680 real templates obtained from 140 fingers. The average matching scores are listed in Table 3, from which it can be seen that the average matching scores reach the maxima when $N$ is around 150.

Table 3 Matching results with DB2_A

| N | 1ST | 2ND | 3RD | 4TH | 5TH | 6TH |
|---|---|---|---|---|---|---|
| 50 | 48.6 | 66.5 | 77.9 | 82.6 | 87.8 | 92.3 |
| 75 | 73.7 | 98.9 | 112.2 | 120.9 | 129.2 | 134.3 |
| 100 | 93.0 | 122.2 | 133.9 | 151.4 | 154.3 | 162.9 |
| 125 | 106.7 | 138.6 | 157.5 | 168.9 | 177.7 | 184.2 |
| 150 | 116.6 | 149.4 | 167.9 | 177.8 | 184.9 | 188.1 |
| 200 | 128.7 | 155.0 | 166.0 | 170.2 | 170.1 | 170.9 |
| 300 | 128.7 | 138.8 | 138.5 | 136.7 | 135.0 | 133.6 |
| 400 | 116.9 | 116.6 | 115.8 | 114.4 | 112.0 | 108.9 |
| 500 | 105.8 | 103.8 | 101.3 | 98.7 | 96.4 | 95.5 |
| 600 | 94.9 | 91.4 | 88.8 | 88.4 | 85.7 | 83.6 |
| 800 | 79.8 | 78.1 | 75.9 | 74.0 | 72.2 | 71.1 |
| 1000 | 69.9 | 67.7 | 66.6 | 64.5 | 62.2 | 60.7 |
| 1200 | 63.0 | 60.8 | 58.5 | 56.5 | 54.6 | 53.4 |
| 2000 | 43.8 | 43.6 | 41.5 | 39.7 | 38.0 | 36.0 |
| 3000 | 35.8 | 34.1 | 31.5 | 30.1 | 28.2 | 26.6 |
| 4000 | 29.9 | 28.3 | 26.0 | 24.5 | 23.0 | 21.2 |

Since the average matching score of RT is ~45.8. Therefore, VT perform better than RT when $N$ is in [50, 1200].

Table 4 gives the matching results between RT and VT. The low scores indicate that VT are significantly different from RT.

Table 4 RT vs. VT (N=200)

|     | AVG  | STD  |
| --- | ---- | ---- |
| 1ST | 1.43 | 1.62 |
| 2ND | 1.84 | 1.63 |
| 3RD | 2.11 | 1.63 |
| 4TH | 2.24 | 1.61 |
| 5TH | 2.32 | 1.60 |
| 6TH | 2.39 | 1.60 |

### 4.3 DBII from PolyU

The DBII contains 1480 real templates obtained from 148 fingers. One synthetic template is paired with each of the 10 real templates pertaining to a finger. For each pair, six different verification templates are constructed with the $1^{ST}$, $2^{ND}$, $3^{RD}$, $4^{TH}$, $5^{TH}$, and $6^{TH}$ nearest neighbor, respectively. The results are plotted in Fig. 3.

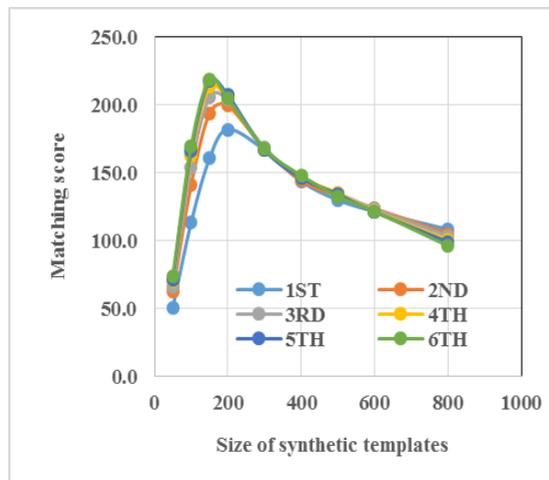

Fig. 3 Average matching scores with DBII

From Fig. 3, we can see that the results of DBII are similar to those of DB1 (4.1) and DB2_A (4.2). Therefore, a common conclusion can be drawn that the proposed approach can achieve equivalent or better performance with proper selection of $N$ and $L$.

In order to answer the question whether the proposed approach can survive multi-generation transformation, we utilize the $1^{ST}$ generation verification templates constructed with $N=1000$ and $L=6^{TH}$ as the real templates to construct the $2^{nd}$ generation templates. For the $2^{nd}$ generation verification templates, four different nearest neighbors are selected: $1^{ST}$, $6^{TH}$, $11^{TH}$, and $16^{TH}$. The matching results are summarized in Table 5.

Table 5 Matching of the $2^{nd}$ generation VT

| N | 1ST | 6TH | 11TH | 16TH |
|---|---|---|---|---|
| 100 | 124.0 | 158.0 | 161.4 | 166.9 |
| 125 | 153.9 | 187.5 | 190.1 | 192.8 |
| 150 | 169.8 | 201.8 | 204.5 | 202.4 |
| 175 | 185.3 | 208.1 | 208.1 | 205.3 |
| 200 | 186.3 | 201.8 | 199.2 | 200.5 |
| 250 | 180.5 | 182.8 | 183.6 | 182.9 |
| 500 | 137.8 | 138.9 | 136.9 | 135.9 |
| 1000 | 116.6 | 113.2 | 115.3 | 113.0 |
| 2000 | 101.9 | 98.3 | 98.6 | 97.1 |
| 4000 | 90.5 | 88.8 | 88.1 | 87.7 |
| 8000 | 84.5 | 83.7 | 83.3 | 82.3 |
| 16000 | 81.8 | 81.5 | 80.7 | 80.2 |
| 32000 | 80.3 | 79.9 | 80.3 | 79.8 |
| 64000 | 79.6 | 79.5 | 79.2 | 79.5 |

From Table 5 it can be seen that the average matching scores of VT are higher than that of RT (71.3) when $N$ is in [100, 64000]. Maximal scores are generated when $N$ is around 175.

The matching results of VT for both generations are given in Fig. 4, from which it can be seen that the $2^{ND}$ generation templates perform better than the $1^{ST}$ generation templates when $N >500$. With $N >2000$, the ranking of the average matching scores is:

$2^{ND}$ generation templates >Real templates >$1^{ST}$ generation templates

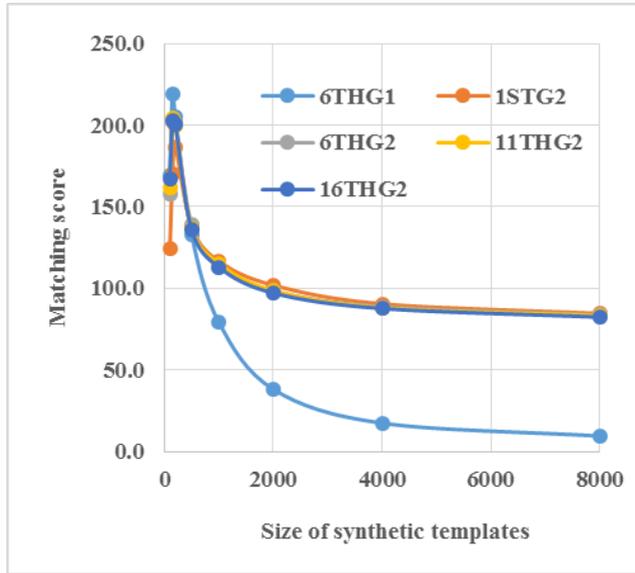

Fig. 4 Effects of $N$ on two generations

The cross generation matching results are given in Table 6, from which we can see that the cross matching score increases as $N$ increases. To avoid possible cross matching (i.e., to reduce the possibility of reversing VT), it is necessary to select $N < 4000$ and $L \geq 6$.

Table 6 Cross generation matching results*

| N | p6c1 | p6c6 | p6c11 | p6c16 |
|---|---|---|---|---|
| 100 | 0.67 | 0.95 | 0.96 | 0.96 |
| 125 | 0.98 | 1.35 | 1.40 | 1.42 |
| 150 | 1.36 | 1.74 | 1.64 | 1.77 |
| 175 | 1.68 | 2.06 | 2.04 | 2.04 |
| 200 | 1.94 | 2.30 | 2.29 | 2.33 |
| 250 | 2.34 | 2.58 | 2.63 | 2.70 |
| 500 | 3.29 | 3.35 | 3.39 | 3.38 |
| 1000 | 4.49 | 3.73 | 3.77 | 3.79 |
| 2000 | 8.76 | 4.03 | 3.98 | 4.01 |
| 4000 | 23.25 | 4.47 | 4.29 | 4.17 |
| 8000 | 57.55 | 5.34 | 4.66 | 4.34 |
| 16000 | 114.85 | 9.79 | 5.98 | 5.00 |
| 32000 | 191.88 | 23.35 | 11.15 | 7.51 |
| 64000 | 287.29 | 58.18 | 26.32 | 16.89 |

*p6 is the $1^{ST}$ generation template, c1, c6, c11 and c16 are the $2^{ND}$ generation templates.

The matching results between the 2$^{ND}$ generation sibling templates are plotted in Fig. 5.

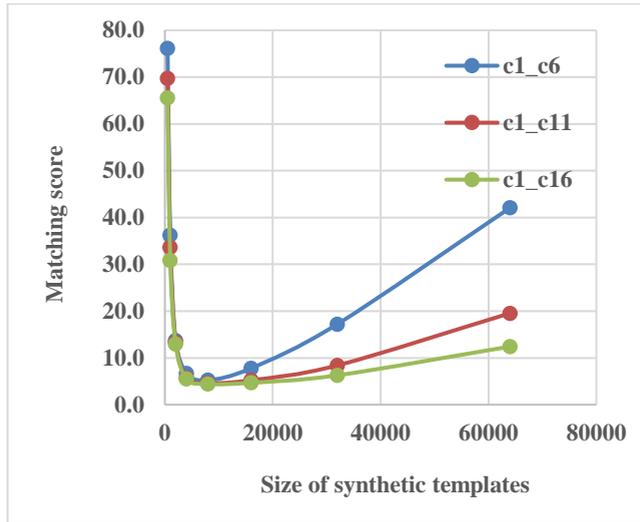

Fig. 5 Inter-sibling matching results

Based the results given in Fig. 5, we believe that cautions must be taken if one chooses to construct the cancelable templates just by changing *L*. To be on the safe side, both ST and *L* should be changed when constructing new VT for a user.

## 5. Conclusion

In this paper a novel approach of constructing cancelable fingerprint template is proposed. Verification template (VT) is constructed by minutiae-wise connecting a real template (RT) to a PRNG generated synthetic template (ST) using the k-nearest neighbor method. VT is a sub-template of ST and satisfies the requirements of cancelable biometrics:
• Diversity: distinctive templates can be generated by using different ST.
• Revocability: a compromised template can be replaced by changing ST.
• Non-reversibility: it is infeasible to recover the original RT given VT because the VT only contains random minutiae.
• Accuracy: matching with VT does not lower the recognition accuracy when *N* and *L* are chosen properly.

According to the results obtained with three publicly available databases, matching between two different VT originated from a same finger produces the maximal score when the ST contain 150 ~200 minutiae.

The proposed approach can be utilized to transform RT across multiple generations. With proper selection of $N$, $2^{ND}$ generation VT can generate higher matching score than RT and $1^{ST}$ generation VT.

Cross-generation matching (parent-child matching) can be avoided by restricting $N$ (ex., $N<8000$) and/or by using a large $L$ (ex., $L= 16$). Cross-sibling matching (child-child matching) is possible. Caution must be taken when constructing multiple cancelable templates just by changing $L$ (i.e., without changing the ST.).

For better security, both the synthetic template and the ordinal number $L$ should be changed when constructing a new template.

In this paper we focused on k-nearest neighbor based minutiae mapping method. Future research will be carried out on finding other mapping methods.

**Acknowledgements**
The author wishes to thank Dr. Javier Ortega-Garcia from Universidad Autonoma de Madrid for allowing us to use the DB2_A database from FVC2006, and to thank Dr. Lei Zhang from Hong Kong Polytechnic University for allowing us to use the PolyU HRF Database.